# Benchmarking Jetson Edge Devices with an End-to-end Video-based Anomaly Detection System

Hoang V. Pham[1]; Thinh G. Tran[1]; Chuong D. Le[1]; An D. Le[2]; Hien B. Vo[1]

[1] Vietnamese German University, Binh Duong 820000, Vietnam
[2] University of California San Diego, La Jolla 92037, USA
`lncs@springer.com`

**Abstract.** Innovative enhancement in embedded system platforms, specifically hardware accelerations, significantly influence the application of deep learning in real-world scenarios. These innovations translate human labor efforts into automated intelligent systems employed in various areas such as autonomous driving, robotics, Internet-of-Things (IoT), and numerous other impactful applications. NVIDIA's Jetson platform is one of the pioneers in offering optimal performance regarding energy efficiency, desirable accuracy, and throughput in the execution of deep learning algorithms. Previously, most benchmarking analysis was based on 2D images with a single deep learning model for each comparison result. In this paper, we implement an end-to-end video-based crime-scene anomaly detection system inputting from surveillance videos and the system is deployed and completely operates on multiple Jetson edge devices (Nano, AGX Xavier, Orin Nano) for benchmarking purposes. The comparison analysis includes the integration of Torch-TensorRT as a software developer kit from NVIDIA for the model performance optimization. The system is built based on the PySlowfast open-source project from Facebook as the coding template. The end-to-end system process comprises the videos collection from camera, data preprocessing pipeline, feature extractor and the anomaly detection. We also provide the experience of an AI-based system deployment on various Jetson Edge devices with Docker technology. Regarding anomaly detectors, a weakly supervised video-based deep learning model called Robust Temporal Feature Magnitude Learning (RTFM) is applied in the system. The approach system reaches 47.56 frames per second (FPS) inference speed on a Jetson edge device with only 3.11 GB RAM usage total. We also discover the promising Jetson device that the AI system achieves 15% better performance than the previous version of Jetson devices while consuming 50% less energy power.

## 1 Introduction

The idea of Smart Cities impacts authority planners and researchers, prompting them to enhance resident safety and security. As the need for residential security grows, there is an increased demand for deploying surveillance cameras in almost every corner to utilize for video analysis. One primary challenge in examining surveillance footage is spotting abnormal events, often requiring extensive human involvement. Many of these intelligent surveillance applications produce large amounts of contextual data, considerable storage, and computing resources.

An Internet-of-Things (IoT) structure enables high-quality adaptability and scalability to meet the growing surveillance needs. Incorporating intelligent surveillance into the IoT framework requires executing computer vision tasks such as vehicle tracking or irregular event detection at the sensor layer of the IoT structure to maintain response time, precision, and energy efficiency. In some critical applications to human safety, like crime-related anomaly detections, response time is a crucial factor. The traditional approach requires uploading large amounts of visual surveillance data to the cloud for analysis, leading to extensive bandwidth and inevitable network latency. Consequently, it prevents the real-time response for the targeting events. On the other hand, the video analysis tasks at the sensor layer of an IoT structure accelerate the capture and process of visual data and make decisions in real-time. It addresses the application issue by focusing only on a short recording clip containing proof of abnormal events, without the need to upload and store intermediate surveillance videos to cloud computing.

As the use of deep learning inference applications on embedded devices continues to grow, the devices are increasingly equipped with hardware accelerators in conjunction with a multi-core CPU and GPU. For instance, the NVIDIA Jetson AGX Xavier board includes a CPU, a GPU, and deep learning accelerators (DLAs). A DLA is also known as a neural processing unit (NPU) for enhancing the operations of deep neural networks. To operate a trained neural network on an embedded device, the utilization of the software development kit (SDK) is a necessary component. The contributions of this paper are summarized as follows:

1) We apply and fine-tune a weakly supervised anomaly detection technique known as RTFM [1] on UCF-Crime and VNAnomaly datasets at video level.
2) The system is built based on an open-source PySlowfast [2] coding template to utilize many beneficial functionalities such as design patterns, efficient data loader for videos, multiprocessing, and multithreading.
3) We utilize TensorRT [3] as the SDK for NVIDIA devices to optimize deep learning applications and improve efficiency in this work.
4) We successfully tested, deployed, and compared an end-to-end anomaly detection system on 3 different types of Jetson edge devices: Jetson Nano, Jetson AGX Xavier, and Jetson Orin Nano.

This work is structured as follows. In section 2, related works are reviewed. In section 3, we discuss about the target edge devices, frameworks and anomaly detection models being used in this system. In section 4, the system workflow and prototype are presented and explained. The data preprocessing and training configuration of anomaly detection models, the deployment's experience on different Jetson Edge devices are specified in section 5. The comparison result of an end-to-end anomaly detection system being deployed on 3 different Jetson devices is described in section 6. Conclusion and future works are drawn in the last section.

## 2    Related Works

Currently, most video surveillance systems primarily emphasize repositories of footage, requiring huge storage and processing capacities. Only a few systems are real-time intelligent surveillance capable of performing object detection, tracking, or anomaly detection tasks, utilizing networking servers and cloud operations. Over the past decade, several approaches have been developed for automated surveillance or object tracking using edge devices. However, only a few focus on an end-to-end anomaly detection system by video-based surveillance.

**Anomaly Detection in Residential Video Surveillance on Edge Devices in IoT Framework**. Mayur R. Parate [4] introduced a development of a framework aimed at anomaly detection within residential video surveillance. The author does the performance analysis on real-world residential surveillance video streams for CPU-only edge devices to gauge the effectiveness of the anomaly detector. Alerts for identified anomalies were disseminated through a Long Range Radio (LoRa) network, ensuring the privacy in the residential setting. Object detection, feature encoding, and trajectory correlations were conducted in Mayur R. Parate's works to identify unusual human behavior in surveillance scenes. Unfortunately, trajectory-based techniques consider only the visual pattern and ignore the importance of targets in complex situations such as crowded scenes, which leads to limited performance [5].

**DeepEdgeBench: Benchmarking Deep Neural Networks on Edge Devices.** Stephan Patrick Baller [6] introduces and evaluates the performance of four Systems on Chips (SoCs): Asus Tinker Edge R, Raspberry Pi 4, Google Coral Dev Board, Nvidia Jetson Nano, along with a microcontroller: Arduino Nano 33 BLE. The evaluation focus on inference time and power consumption. Authors also offer a technique for measuring the devices' power consumption, inference time, and accuracy. The findings reveal that, concerning a Tensorflow-based quantized model, the Google Coral Dev Board outperforms the others in inference time and power consumption.

**Benchmark Analysis of Jetson TX2, Jetson Nano and Raspberry PI using Deep-CNN.** Ahmet Ali Süzen [21] examines and compares the performance of single-board computers, namely NVIDIA Jetson Nano, NVIDIA Jetson TX2, and Raspberry PI4, by employing a Convolutional Neural Network (CNN) algorithm trained on a fashion product images dataset. A 2D CNN model was constructed to categorize 13 distinct fashion products, utilizing a dataset composed of 45K images. Performance indices were established based on resource consumption (GPU, CPU, RAM, Power), accuracy, and cost. To delve into the performance variations of the single-board computers, the dataset was divided into partitions of 5K, 10K, 20K, 30K, and 45K for model training and testing. The principal objective is to achieve optimal accuracy using minimal hardware resources in deep learning applications. However, the deep learning model used in this paper is a light-weight custom CNN model, and the comparison result might not be generalized for different models.

**Benchmarking Jetson Platform for 3D Point-Cloud and Hyper-Spectral Image Classification**. Shan Ullah [20] presents a performance benchmarking of Jetson platforms (Nano, TX1, and Xavier) by utilizing deep learning algorithms namely PointNet. The paper explores the impact of algorithm optimization and hardware acceleration by deploying a range of dense deep learning architectures across the three Jetson platforms. Two contrastingly different data types were used for this purpose: the ModelNet-40 data-set and hyperspectral image datasets. The objective was to test the capabilities of GPU-heavy TensorFlowgpu code through 3D data and challenge the CPU cores with Theano-based code through Hyperspectral images (HSI) data. The assessment parameters considered for the evaluation included inference time, the maximum number of simultaneous processes, resource utilization per process, and efficiency.

## 3 Target edge devices, Frameworks, and Anomaly Detection model

This section introduces different Jetson edge device architectures and the hardware overview for system deployment comparison. We also present the system frameworks, the optimization techniques, and the foundation architecture information of the chosen anomaly detection model.

### 3.1 Target Edge Devices

**Nvidia Jetson Nano**. The Jetson Nano [7] is an edge computing device offered by Nvidia, featuring their high-performance Graphics Processing Units (GPUs) for accelerated artificial intelligence (AI) processing at the edge. The device is based on the Maxwell microarchitecture and a single streaming multiprocessor (SMs) with 128 CUDA cores, enabling simultaneous execution of multiple neural networks. As the entry-level model in the Jetson family, the Nano delivers a peak performance of 4 Tera Operations Per Second (TOPs), with available memory configurations of 2 GB or 4 GB. Our experiments utilized the 4GB RAM and 10W power mode variant to maximize performance.

**Nvidia Jetson AGX Xavier**. Jetson AGX Xavier [8] is another edge computing platform from NVIDIA, designed for autonomous machines and intelligent edge devices. It is built around the Xavier SoC (System-on-Chip), which combines a high-performance CPU, GPU, and dedicated AI acceleration engines into a single chip. The device is built on an 8-core NVIDIA Carmel Arm 64-bit CPU and 256-core NVIDIA Volta GPU microarchitecture, and 16 SMs with 4 CUDA cores each which equals 256 CUDA cores total. This type is an advanced-level model of the Jetson family, the AGX Xavier delivers a peak performance of 32 TOPs, with available memory configurations of 16GB and 32GB. Our experiments utilized the 32GB RAM and 30W power mode variant to maximize performance.

**Nvidia Jetson Orin Nano**. The Jetson Orin Nano [9] is designed to accelerate entry-level edge AI applications. It sets a new standard for creating entry-level AI-powered robots, smart drones, and intelligent cameras while simplifying the process of starting with the Jetson Orin Nano series. It is available in 4GB and 8GB versions. The 8GB version has a 1024-core Ampere architecture GPU with 32 Tensor cores and a peak clock speed of 625 MHz. The 4GB version has a 512-core Ampere architecture GPU with 16 Tensor cores and a peak clock speed of 625 MHz. Jetson Orin Nano 8GB delivers the maximum performance of 40 TOPs, even higher than those of Jetson AGX Xavier 32GB. Our experiments utilized the 8GB RAM and 15W power mode variant to maximize performance.

Table 1 provides an overview of basic hardware specification of the three Jetson Edge devices that we used to do the benchmarking.

Table 1. Overview of Jetson Edge devices.

|  | **Jetson Nano 4GB** | **Jetson AGX Xavier 32GB** | **Jetson Orin Nano 8GB** |
|---|---|---|---|
| Publish day | October 2020 | September 2018 | January 2023 |
| AI Speed | 4 TOPs | 32 TOPs | 40 TOPs |
| GPU | 128-core NVIDIA Maxwell™ GPU | 512-core NVIDIA Volta™ GPU with 64 Tensor cores | 1024-core NVIDIA Ampere GPU with 32 Tensor Cores |
| CPU | Cortex®-A57 64-bit, 1.43 GHz | 8-core NVIDIA Carmel Arm®v8.2 64-bit, 2.26GHz | 6-core Arm® Cortex®-A78AE v8.2 64-bit, 2.2GHz |
| Memory | 4GB 64-bit LPDDR4 1600MHz frequency | 32GB 64-bit LPDDR4x 2133MHz frequency | 8GB 128-bit LPDDR5 2133MHz frequency |
| Power | 5W - **10W** | 10W - **30W** | 7W - **15W** |

**Camera**. The Logitech C920 [10] is a PRO HD Webcam that delivers full HD 1080p video calling with stereo audio. It has a full HD glass lens, a 78° field of view, and HD auto light correction, which produces bright, contrasted images even in dim settings. The camera is affordable and has excellent HD video quality with reliable autofocus. In this paper, Logitech C920 is the primary USB camera for the end-to-end anomaly detection system on multiple Jetson Edge devices.

### 3.2 Inference, Optimization, and Deployment frameworks

**Pytorch.** Pytorch is an open-source machine learning library developed primarily by Facebook's artificial intelligence research (FAIR) group. It is explicitly built for computer vision and natural language processing tasks, but its flexibility and performance characteristics make it applicable to various other tasks. Pytorch uses Tensors for its operation, similar to NumPy's multidimensional arrays, but with additional capabilities, such as running on GPUs to accelerate computational abilities. Dynamic Computation Graph, or Define-by-Run Graph, is a unique and powerful feature, unlike other popular frameworks like TensorFlow (before Version 2.0) or Theano, which uses the Define-and-Run approach where the computational graph is defined once and run multiple times. Pytorch builds and modifies the graph dynamically during runtime, offering flexibility and debugging benefits. It also supports the Autograd Module to calculate the models' backpropagation without manual implementation automatically. In general, the flexibility and intuitive interface of Pytorch make it one of the most popular frameworks for deep learning research and development.

**Torch-TensorRT**. Torch-TensorRT [3] is an SDK provided by NVIDIA that allows models to run efficiently on NVIDIA hardware, which meets limited resource requirements. It makes models lighter and faster by using optimization techniques as follows. Fig. 1. illustrates how Torch-TensorRT uses the Builder module to generate an optimized inference engine. This is done using specified optimization configuration parameters and a given neural network model. The Builder internally processes the following actions. Firstly, it calibrates the model to run with lower precision (FP16 or

INT8), which can significantly reduce the computation time while maintaining acceptable accuracy. Secondly, it fuses multiple layers and tensors into a single operation, reducing memory bandwidth and latency. Finally, Dynamic Tensor Memory is applied to allocate and release memory for intermediate tensors, reducing memory usage. The workflow then generates the final optimized execution model that performs the inference. The Torch-TensorRT runtime module takes charge of loading the optimized engine. It then deserializes it to establish execution contexts and assigns a Processing Element (PE) to the engine. Overall, Torch-TensorRT dramatically improves the performance of end-to-end anomaly detection systems, enabling us to deploy and run on multiple resource-limited Jetson Edge Devices efficiently.

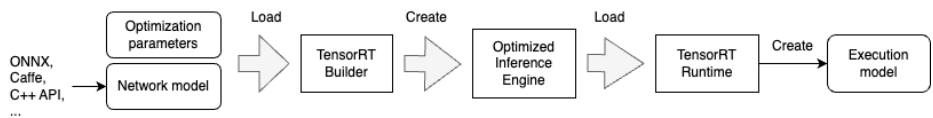

**Fig. 1.** NVIDIA Torch-TensorRT workflows

**Docker.** The benefit of container technology is that it can self-containerize an application and its necessary components, allowing for smoother deployment. This aspect can be effectively accomplished using Dockers [11]. Similar to a Virtual Machine (VM), each Docker operates in isolation, but it doesn't require the support of a guest operating system. This makes it less resource-intensive, and thus it can be easily installed on edge devices without much effort. Fig. 2. shows the workflow of Docker technology when deploying the system on Jetson devices. The users create and encapsulate the application system into a Docker image using the docker engine and Dockerfile, which includes the application source code and associate dependencies installation command in detail. Users then register the created Docker image to the registry server on the internet. The registry serves as a warehouse for images, and when a pull command is run, these images are distributed from the registry to the target devices for deployment. The host runs the encapsulated Docker image on the Jetson target computer and creates a Docker container to run the end-to-end application.

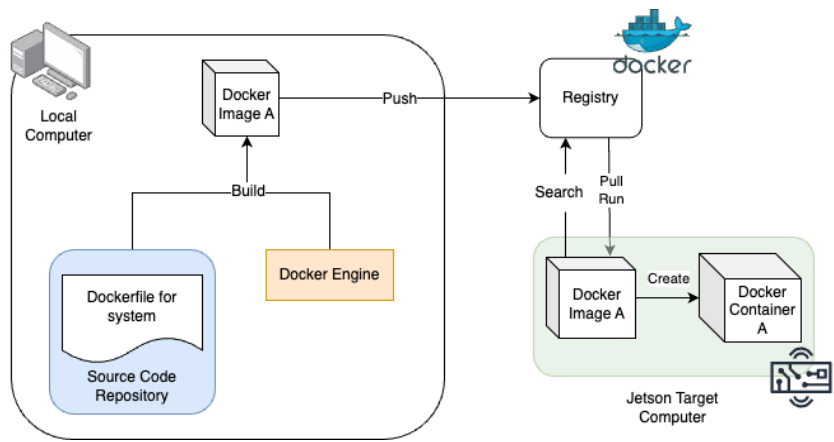

**Fig. 2.** Docker workflows

### 3.3 Feature Extractor and Anomaly Detection model

ResNet50-I3D non-local is a feature extractor model designed explicitly for video analysis tasks. It builds upon the famous ResNet50 architecture, which has been widely used for image classification tasks. The "I3D" block stands for Inflated 3D Convolutions, which is a technique developed by Yukun Huang [16] to improve the performance of convolutional neural networks on time series data like videos. The idea behind this technique is to inflate the spatial dimensions of the input data so that it can capture longer-term dependencies between frames more effectively. ResNet50-I3D non-local takes the output from the ResNet50 model and applies a series of inflated 3D convolutions, followed by pooling layers, to reduce the dimensionality of the feature maps. This allows the model to learn local and global features within each frame and across multiple frames, leading to improved performance on video analysis. The key advantage of ResNet50-I3D non-local compared to other video feature extractors is the ability to handle long-range video dependencies [16]. It mainly results from a " non-local " technique in the chosen feature extractor. It helps to capture long-term patterns and trends over larger distances.

Yu Tian et al. [1] proposed a novel anomaly detection method named Robust Temporal Feature Magnitude (RTFM) learning. Like Multi-Instance Learning (MIL) [12], RTFM learns from weakly labeled videos to identify whether a snippet is normal or abnormal. Each video is divided into a bag of T snippet clips containing a constant number of continuous frames. The algorithm relies on the temporal feature magnitude of these snippets, where snippets with low magnitude represent normal snippets and high magnitude denotes abnormal ones. However, MIL poses some drawbacks, namely: 1) The top anomaly score in abnormal videos may not be from abnormal snippets; 2) Normal snippets randomly chosen from normal videos may be relatively easy to fit, which could pose a challenge to the convergence of the training process; 3) In case, there are more than one abnormal events in a video, we miss an opportunity for one more effective training process, and 4) The use of classification score provides a weak training indicator that does not enable good separation between normal and abnormal snippets. Therefore, Yu Tian et al. utilizes some enhancements to tackle the above issues of MIL in RTFM, as follow 1) By assuming the mean magnitude of abnormal snippets larger than that of normal snippets, the probability of selecting abnormal snippets from abnormal videos increase; 2) The hard normal snippets selected from normal videos will be more challenging to fit, thereby improve the convergence of training process; 3) Enable to include more abnormal snippets per anomaly video; and 4) Recognizing positive instances with feature magnitude is more robust compared to the usage of classification score from MIL methods.

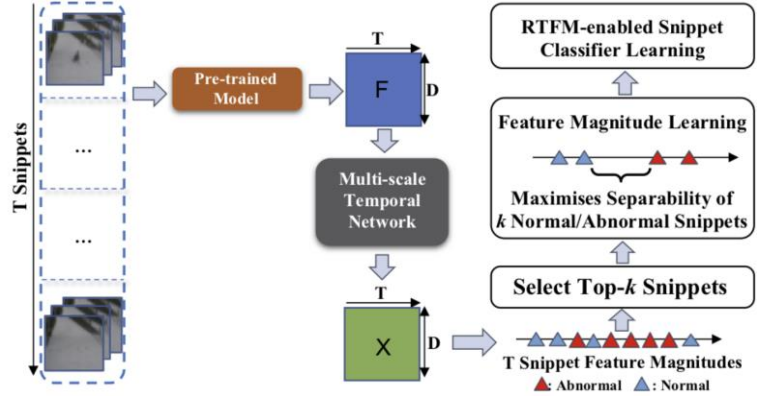

**Fig. 3.** RTFM workflow from [1]

Fig. 3. shows the step-by-step workflow process of the RTFM model. Pre-trained model Resnet-50 I3D first extracted the feature from these snippets. F represents the features with dimension D of T snippets. The Multi-scale Temporal Network (MSTN) [13] incorporates 2 key modules: the Pyramid of Dilated Convolutions (PDC) [14] and Temporal Self-Attention (TSA) [15]. This combination captures multi-resolution local and global temporal dependencies among video snippets. The output of MNT is denoted as X. The process then classified each snippet as normal or abnormal by utilizing l2 normalization to determine feature magnitude. Following this, the method chose k snippets that displayed the highest scores. To maximize the separability of k normal and abnormal snippets, the Feature Magnitude Learning phase minimized the highest feature magnitude of snippets from normal videos while at the same time maximizing the highest feature magnitude of snippets from anomalous videos. Finally, the RTFM used k snippets with the highest feature magnitude to train with its loss function.

## 4   Methodology

The PySlowFast open-source project includes implementations of several models for Video Action Recognition tasks. Using its coding template, we removed all the action recognition components and incorporated the model architecture code of RTFM and the corresponding data processing pipeline. As for Docker, we creat our own Docker image based on the NVIDIA L4T PyTorch [19] image, which is specifically provided with pre-installed Pytorch and torchvision for harnessing the power of Jetson's GPU. Each Jetson Edge device might require a unique L4T Docker image version. The Docker integrates all directories and scripts from the active project into an image and performs initial commands such as updates, upgrades, and pip-python installations.

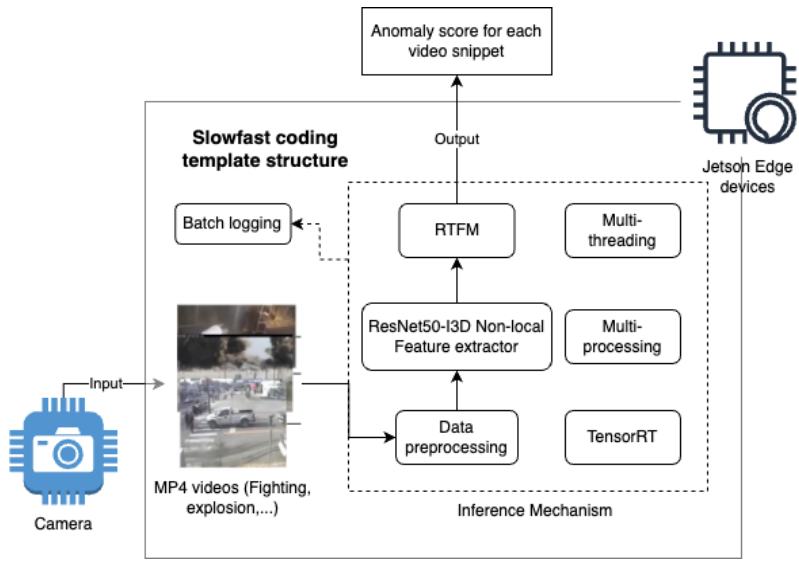

**Fig. 4.** Anomaly Detection system workflow

Fig. 4. illustrates the overall architecture of an end-to-end anomaly detection system. Surveillance videos collected from cameras undergo data preprocessing before they are fed into the ResNet50-I3D non-local feature extractor. Output features are subsequently inserted into the RTFM anomaly detection model, resulting in the final prediction. All the inference steps of the system's workflow are fully operated directly on the Jetson edge devices to maximize the potential advantages of edge computing technology. The system has been designed to benefit from the logging, multiprocessing, multithreading, and video data loader mechanism aspects of the PySlowfast [8] template. The logging mechanism records the system activities during each batch.

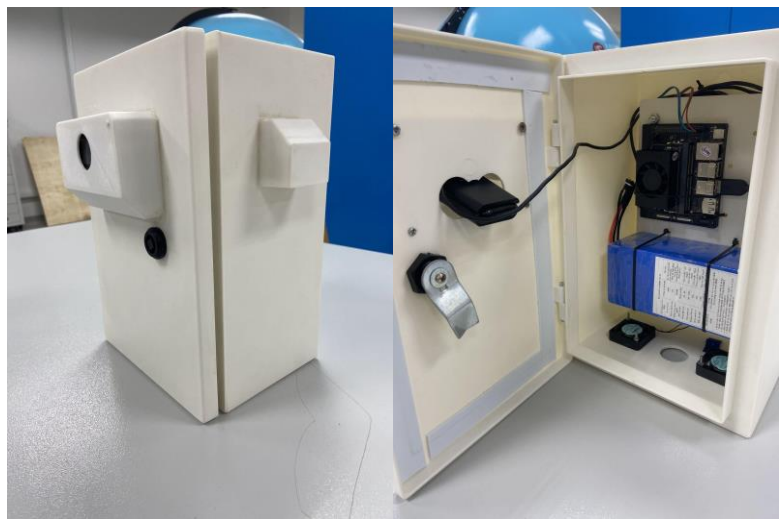

**Fig. 5.** Outside and Inside of Anomaly Detection system prototype

In order to complete a demo of the system, we design and integrate the 3D cover box for the system setup. Fig. 5. visualizes the overview of the outside and the inside of a hardware cover box for the anomaly detection system. The small board on the top right is a single-board computer Jetson Orin Nano. Connecting to the Jetson board on the door of the cover box in the left side is the camera Logitech C920 which collects the surveillance videos and input them directly to the Jetson main system. The blue rectangle block at the bottom of the box is the battery for the remote operation. The surveillance videos are collected from the camera and input directly into the system deployed on Jetson devices to perform the anomaly detection on the field.

## 5   Experiments

### 5.1   Anomaly Detector Training configuration

The UCF-Crime dataset is proposed in "Real-World Anomaly Detection in Surveillance Videos" 2018 by Waqas Sultani [17]. It includes long, untrimmed surveillance videos that represent 13 real-world anomalies such as Abuse, Arrest, Arson, Assault, Road Accident, Burglary, Explosion, Fighting, Robbery, Shooting, Stealing, Shoplifting, and Vandalism. The collection consists of about 1900 normal and abnormal Youtube videos, which stand for 128 hours in total. Another dataset that we trained and performed the evaluation is the UIT-VNAnomaly dataset. The dataset is proposed by Dung T.T. Vo [18] in "UIT-Anomaly: A Modern Vietnamese Video Dataset for Anomaly Detection." It is collected from Vietnam and contains various anomalous behaviors, including Stealing, Traffic Accident, Fighting, Unsupporting Behavior, Against and Dog Thief. The VNAnomaly dataset contains 224 abnormal and normal videos, equal to 3.5 hours. Table 2 provides the type of abnormal cases of each public dataset in detail.

**Table 2.** Fine-tuning result comparison on benchmark datasets

| Dataset | Anomaly labels |
| --- | --- |
| UCF-Crime | Abuse, Arrest, Arson, Assault, Road Accident, Burglary, Explosion, Fighting, Robbery, Shooting, Stealing, Shoplifting, and Vandalism |
| UIT-VNAnomaly | Stealing, Traffic Accident, Fighting, Unsporting Behavior, and Against |

Regarding the data preprocess pipeline, RTFM requires strict step-by-step order of pre-processing techniques and parameter input specification. The pipeline converts the input data into torch Tensor datatype, then resizes those tensors to shape 256. After that, it applies a 10-crop data augmentation technique that involves randomly taking ten different crops of an image with the size 224 and using them as separate inputs to the model. The crops are taken from the four corners, center, and their mirrored versions. Finally, those crop images are stacked together, and the normalization algorithm with provided mean and standard deviation is applied to finalize the data processing pipeline. The mean and standard deviation values should be [114.75, 114.75, 114.75] and [57.375, 57.375, 57.375] respectively, to successfully reproduce the video feature that confirms the result the same as those provided by the authors. Fig. 6. shows the instruction of data preprocessing pipeline for video input before inputting into the feature extractor model.

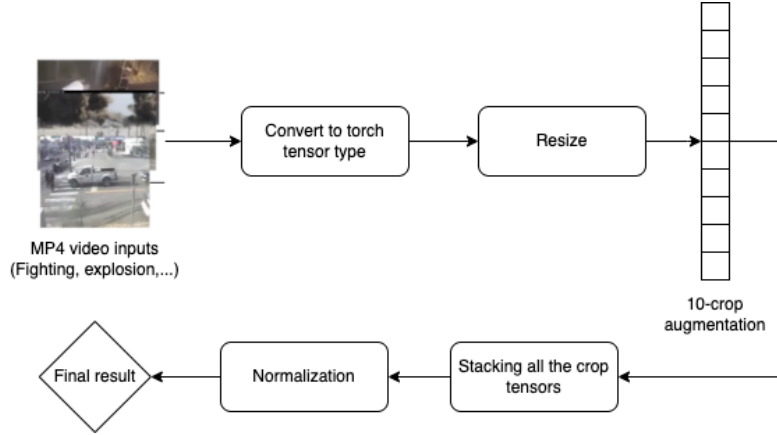

**Fig. 6.** Data preprocessing pipeline

Our system approach breaks each input video into 32 shorter video snippets, each containing 16 sequential frames. We apply the anomaly detection model to both the UCF-Crime and UIT-VNAnomaly datasets. The feature extractor, which uses the result from layer mix_5c of the pre-trained ResNet50-I3D non-local model, initially extracts a 2048D feature from the input video. We retain all layer parameters of the RTFM model as provided in its original form, and the dropout rate in the final fully-connected layer is set to 70%. The fine-tuning of the RTFM model is done based on the pre-trained weight provided by the author using the UCF-Crime dataset. We employ the Adam optimizer for these fine-tuning experiments with a weight decay of 0.005 and a batch size of 16. The learning rate starts at 0.001.

### 5.2 System testing on Jetson Edge devices

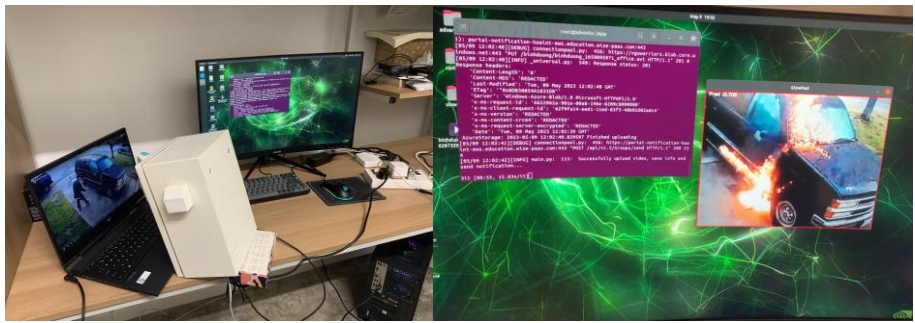

**Fig. 7a (left).** Experiment Setup for demo videos testing; **Fig. 7b (right).** System Process monitor Visualization

The figure 7a on the left depicts the experiment setup to test our anomaly detection system on some demo videos. The figure 7b on the right shows the monitor visualizing the moment when the system detects an abnormal event. Because the crime-related abnormal events rarely appear in the real-world scenarios which poses a challenge to collect them. Therefore, the experiment concept is setted to test the system on some demo videos collected from Youtube. The RTFM model never saw these private videos before. We collect 10 different videos containing crime-scene related events with the types of crime similar to those on the UCF-Crime dataset proposed before. For the demo process, we visualize the system process on the monitor while the system can operate solely on Jetson devices in real scenarios. The terminal on figure 7a contains the system logs that tracks all the events happening during the system running. The logs mechanism is provided by the PySlowfast codebase template of the system. In figure 7b, a small window view visualizes the view from the camera in real-time. On the top left of the window, a small number represents an abnormal score for each snippet and a red line covering the edge of the window appears when the score surpasses the provided abnormal threshold.

### 5.3 System Deployment on Jetson Edge devices

The optimal performance of the anomaly detection system requires Jetson devices to operate at maximum capacity. We thus set all the Jetson devices to their highest mode and activate the Jetson clocks. The Pytorch version used in this system is v1.11, which offers stability and broad compatibility with all the required dependencies. However, one significant challenge in deployment is the installation of Torch-TensorRT on different types of Jetson Edge devices. Torch-TensorRT is built based on Bazel [8], which handles extensive codebases with multiple dependencies and incorporates a built-in interface for the library in Jetson devices. It's important to note that each series of Jetson might require different versions of Torch-TensorRT.

**Table 3.** System dependencies requirement on Jetson devices.

|  | NVIDIA L4T PyTorch version | Torch-TensorRT version | Python version |
|---|---|---|---|
| Deployment on Jetson Nano | Jetpack 4.6 (PyTorch v1.8.0) | v1.0.0 (Plus additional script from v1.1.1) | 3.6 |
| Deployment on Jetson AGX Xavier | Jetpack 5.0.0 (PyTorch v1.11.0) | v1.1.1 (Plus additional script from v1.3.0) | 3.8 |
| **Deployment on Jetson Orin Nano** | **Jetpack 5.1.1 (PyTorch v1.11.0)** | **v.1.1.1** | **3.8** |

Jetson Nano is compatible with Torch-TensorRT version v1.0.0. However, this version only partially supports functionalities of the chosen anomaly detection model. To resolve this, we should examine the functionalities in Torch-TensorRT version v1.1.1, specifically looking at the scripts that are not present in v1.0.0, and manually add them to the Torch-TensorRT source v1.0.0 to ensure the successful system's running. In contrast, deploying on Jetson AGX Xavier presents unique challenges compared to

Jetson Nano. AGX Xavier supports the NVIDIA L4T PyTorch version 5.0.0, which is incompatible with Torch-TensorRT version v1.1.1, while v1.1.1 is required for AGX Xavier. This issue can be addressed by incorporating selected scripts from Torch-TensorRT v1.3.0 into Torch-TensorRT version v1.1.1, which is compatible in this case, to enable the support for L4T Docker version 5.0.0. Lastly, with Jetson Orin Nano, the deployment process is comparatively smooth due to compatibility with version 5.1.1 of the L4T Pytorch Docker and Torch-TensorRT version v1.1.1. The fortunate match of compatible versions in Orin Nano reduces the conflict between the RTFM support components and permits more seamless system integration. The table 3 depicts the overall dependencies requirement for the system to run successfully on each target Jetson device.

In terms of Pytorch verison, NVIDIA L4T PyTorch images do not provide suitable PyTorch version to completely run the anomaly detection system. For example in Jetson AGX Xavier, L4T image supports Pytorch version v1.12.0, while the system specifically requires version v1.11.0 instead. Therefore, we should reinstall the Pytorch version to confirm the successful operation. Regarding the camera connection, the AI system requires a link to a USB camera. Successfully integrating USB cameras with Docker containers on Jetson Edge devices involves a fair amount of necessary adjustments. In this context, we need to enable the Docker container to access the camera using the xhost access control program in Linux. When executing the command to run Docker images, the camera devices should be identified and declared in the X11 volume environment. With regard to system architecture, our system's Docker image is built from a local computer having an x86 architecture, and we have to make to image work in the Jetson architecture, which is arm/v8. Docker's "docker buildx build" feature was especially helpful here, as it allowed us to construct the images based on the architecture of the target computer.

## 6   Result and Discussion

### 6.1   Anomaly detection

Table 4. Fine-tuning result comparison on benchmark datasets

| RTFM | AUC (%) |
|---|---|
| Trained on UCF-Crime dataset | 84.39 |
| Trained on UIT-VNAnomaly dataset | 88.4 |

Table 4 presents the fine-tuning outcome of the RTFM model applied on two datasets: UCF-Crime and UIT-VNAnomaly, using AUC as the evaluation metric. The pre-trained RTFM fine-tuning on the UCF-Crime dataset can still nearly keep the same result as the one provided by the author with nearly 84.39% AUC. It proves the accurate implementation of the data preprocessing pipeline and the architecture of the feature extractor. The fine-tuned RTFM model on the UIT-VNAnomaly dataset still reaches 88.4% AUC accuracy as well.

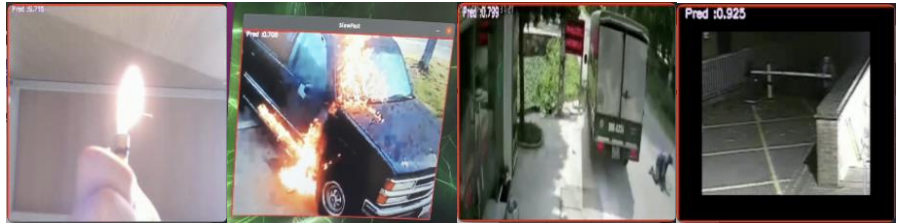

**Fig. 8.** Sample detected frames from demo videos

We choose 10 videos containing crime-scene abnormal events from Youtube and from real-world scenarios that the system might have never seen before for the demo testing. The system correctly detects 9 out of 10 videos and specifically points out the targeted frames sequence. Figure 8 illustrates some captured frames when the system detects anomaly actions from the video demos. The first two frames from the left represent the arson case when we turned on the lighter and someone burned the cars. The third frame is for the accident case when the pedestrian is hit by the truck and the fourth is for the vandalism case when someone destroys the traffic barrier. The small prediction score appearing on the top left of each frame represents the abnormal score of the associate clip snippet. The higher the prediction score, the higher chance the events currently happening in the surveillance video are abnormal. We set the threshold for the prediction score to be 0.7.

### 6.2 Benchmark Jetson devices for end-to-end anomaly detection system

The table 5 provides a comparative analysis of the end-to-end anomaly detection system across three distinct Jetson Edge Devices: Jetson Nano, Jetson AGX Xavier, and Jetson Orin Nano. The evaluation starts from the camera collecting surveillance mp4 videos, data processing pipeline, the feature extractor to the anomaly detection RTFM model. The system contains two deep neural network models. The RTFM has 24.719M parameters with 3.461 Giga Floating Point Operations Per Second (GFLOPs) and 34.582M parameters with 38.272 GFLOPs for feature extractor ResNet50-I3D Non-Local. Adding them together, the end-to-end system containing nearly 59.301M parameters and 41.733 GFLOPs operations is evaluated on multiple Jetson devices as in the table 5. The metrics considered for this comparative assessment are Random Access Memory (RAM) consumption and Frames Per Second (FPS). Figure 9a and 9b are the visualizations of end-to-end system evaluation on RAM and FPS respectively.

**Table 5.** End-to-end system evaluation on multiple Jetson Edge devices.

|  | Jetson Nano (4GB) | | Jetson AGX Xavier (32GB) | | Jetson Orin Nano (8GB) | |
|---|---|---|---|---|---|---|
|  | Without Torch-TensorRT | Torch-TensorRT | Without Torch-TensorRT | Torch-TensorRT | Without Torch-TensorRT | Torch-TensorRT |
| RAM usage (GB) | X | 2.61 | 5.72 | 3.74 | 4.94 | **3.11** |

| Frame per Second (FPS) | X | 1.55 | 29.57 | 41.65 | 36.02 | **47.56** |

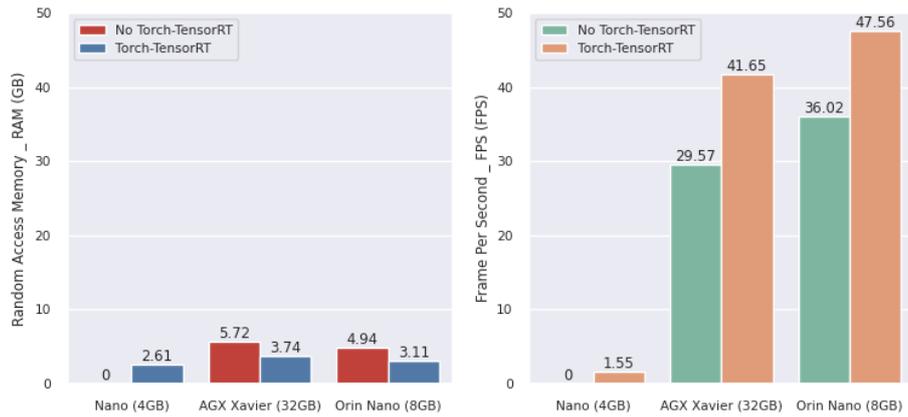

**Fig. 9a (left).** End-to-end system comparison by RAM usage (GB); **Fig. 9b (right).** End-to-end system comparison by FPS

In this testing environment, all devices shared the same setup parameters. The system needs 3.11 GB RAM for active operation, in contrast to the Jetson AGX Xavier's 3.74 GB consumption, denoting a more RAM-efficient usage. The system is only operational on Jetson Nano with Torch-TensorRT because of the limited RAM resource compared to the required RAM usage from the system. The Jetson Nano with Torch-TensorRT displayed minimal RAM consumption at 2.61 GB while having the slowest speed, 1.55 FPS only. Contrastingly, the Jetson Orin Nano proved the fastest speed with 47.56 FPS, amounting to almost 30 times the speed of the Jetson Nano, additionally surpassing the Jetson AGX Xavier by nearly 15% under identical setup conditions. During anomaly detection system operation, the Jetson Orin Nano exhibited half the power consumption compared to Jetson AGX Xavier. As such, the Jetson Orin Nano 8GB RAM with Torch-TensorRT emerged as the most effective device for anomaly detection system implementation, surpassing the other compared options in terms of efficiency.

## 7 Conclusion

The article introduces the comparative analysis of an end-to-end anomaly detection system on 3 different Jetson Edge Devices: Jetson Nano, Jetson AGX Xavier, and Jetson Orin Nano. The comparison result includes the application of an optimized framework Torch-TensorRT into the system. We also provide the system deployment experience on various Jetson Edge devices using Docker for containerization. Noticeably, the performance result of the anomaly detection system on Jetson Orin

Nano 8GB is 15% faster speed and 20% less memory usage compared to Jetson AGX Xavier under the same setup configuration. At the same time, Jetson Orin Nano consumes only half the amount of energy than AGX Xavier. With regard to future work perspectives, we can consider some directions as below:

1) The connection between the system and cloud technology to construct a complete surveillance solution system for anomaly detection purpose.
2) An application of this end-to-end system in different real-world surveillance scenarios and the analysis on device placements.
3) Testing the robustness and efficiency of the system under different video quality, resolution and environmental conditions.
4) More detail study about energy efficiency, data privacy and security for the surveillance video collected on the field.

## Acknowledgement

Grateful for Mr. Dat Thanh Vo, from University of Windsor, Canada, for the support on the process of designing a 3D cover box.
Grateful for Mr. Anh Duy Pham, from Hochschule Bonn-Rhein-Sieg Sankt Augustin, Germany, for the constructive comments during the deployment process.